\pdfoutput=1

\documentclass[11pt]{article}

\usepackage[]{acl}

\usepackage{times}
\usepackage{latexsym}

\usepackage[T1]{fontenc}

\usepackage[utf8]{inputenc}

\usepackage{microtype}

\usepackage{inconsolata}

\usepackage{graphicx}
\usepackage{array} 
\usepackage{tabularx}

\usepackage{booktabs}
\usepackage{multirow}

\usepackage{pifont}  
\newcommand{\cmark}{\ding{51}}%
\newcommand{\xmark}{\ding{55}}%

\usepackage{framed,xcolor}

\usepackage[]{fancyhdr}

\renewcommand{\sectionmark}[1]{}

\usepackage{tabu}

\newcommand\tag[1]{\textbf{#1}\hspace{3pt}}

\title{AustroTox: A Dataset for Target-Based Austrian German\\ Offensive Language Detection}

\author{Pia Pachinger \\ TU Wien \\ pia.pachinger@tuwien.ac.at \And
        Janis Goldzycher \\ University of Zurich \And
        Anna Maria Planitzer \\ University of Vienna
        \AND
        Wojciech Kusa \\ TU Wien \And
        Allan Hanbury \\ TU Wien \And
        Julia Neidhardt \\ TU Wien}

\begin{document}
\maketitle

\thispagestyle{fancy}

\begin{abstract}
Model interpretability 
in toxicity detection greatly profits from token-level annotations. However, currently such annotations are only available in English. We introduce a dataset annotated for offensive language detection sourced from a news forum, notable for its incorporation of the Austrian German dialect, comprising 4,562 user comments. In addition to binary offensiveness classification, we identify spans within each comment constituting vulgar language or representing targets of offensive statements. We evaluate fine-tuned language models as well as large language models in a zero- and few-shot fashion. 
The results indicate that while fine-tuned models excel in detecting linguistic peculiarities such as vulgar dialect, large language models demonstrate superior performance in detecting offensiveness in AustroTox. We publish the data and code\footnote{\url{https://www.pia.wien/austrotox/},\\ \url{https://web.ds-ifs.tuwien.ac.at/austrotox/}}.
\end{abstract}
\small
\textcolor{red}{\textbf{Content warning}: This paper contains examples of offensive language to describe the annotation scheme.}
\normalsize

\section{Introduction}

In recent years, research in the domain of content moderation has transitioned from a unidimensional to a multidimensional perspective \citep{cabitza2023toward}. A one-size-fits-all approach is unable to  accommodate the diverse needs of global users \citep{cresci2022personalized} whose perceptions of what constitutes harmful content is contingent upon individual, contextual, and geographical factors \citep{bormann2022perceptions, jiang2021understanding, kumpel2023differential}. Scholars call for less centralized and more personalized mechanisms of moderation to account for such multifaceted differences \citep{jhaver2023personalizing}, particularly when it comes to country-specific and subsequently linguistic nuances \citep{jiang2021understanding, demus2023automatische}. 
Intolerant user comments (e.g., offensive stereotyping), in contrast to incivil user comments (e.g., vulgarity), are perceived as more offensive and as a stronger threat to democratic values and society, as well as receiving a stronger support for deletion \citep{kumpel2023differential}. This highlights the importance of moderation approaches that include a more nuanced understanding of online norm violations. 
\begin{figure}[]
\includegraphics[width=\linewidth]{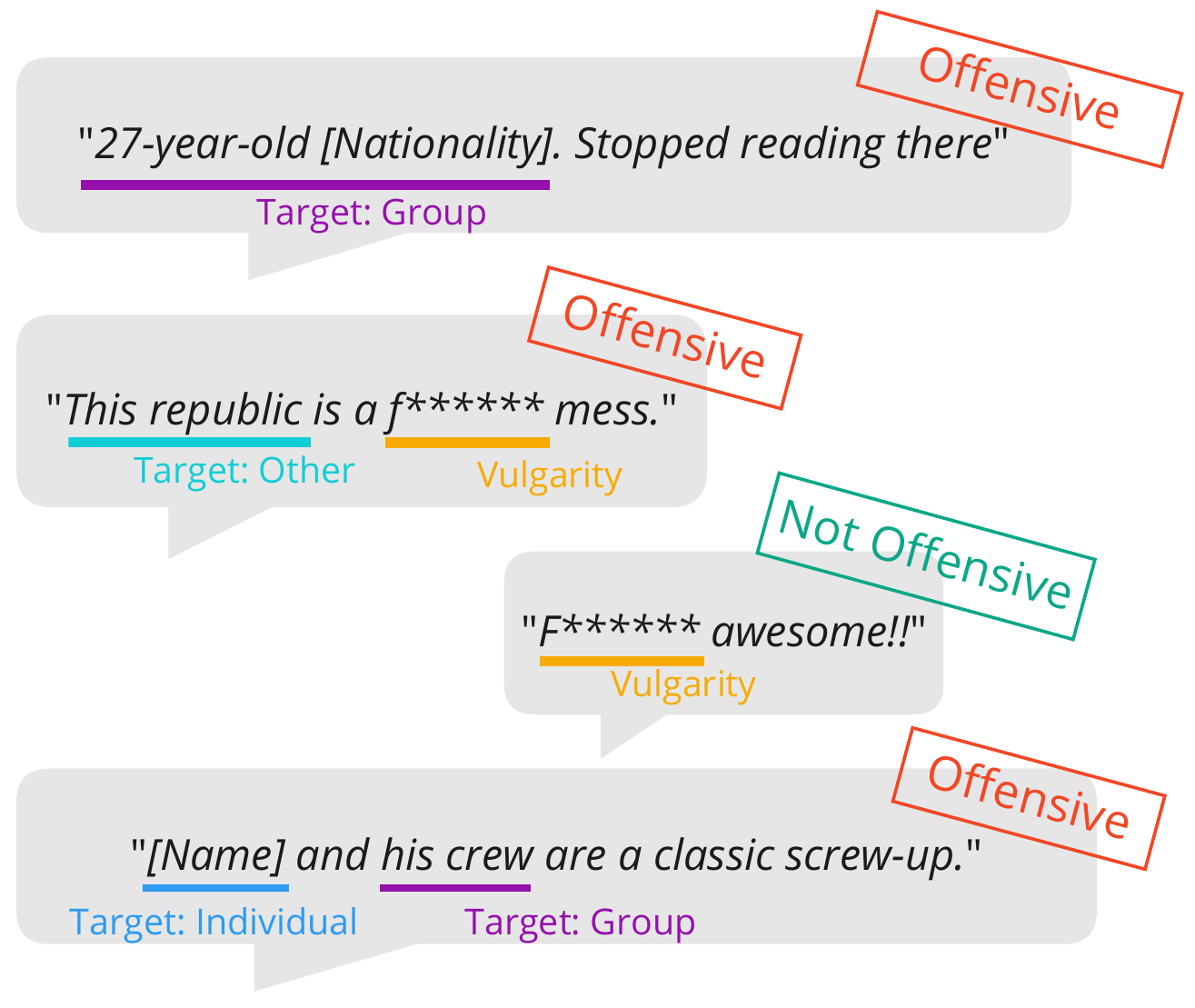}
\caption{
The posts show the importance of the target of an offensive statement in determining its severity and demonstrate how non-offensive remarks can be vulgar.
}
\end{figure}
\label{fig:examples}
\begin{table*}[ht]
    \centering
    \begin{tabularx}{\linewidth}{X l r c l}
        \toprule
        \textbf{Dataset} & \textbf{Source} & \textbf{\#Posts} & 
        \textbf{Spans} & \textbf{Selected Annotations}\\
        \midrule
        \citeauthor{bretschneider2017detecting} & FB & 5,600 & \xmark & Moderate HS, Clear HS \\
        One Million Posts (\citeyear{schabus2017one}) & DerStandard & 11,773 & \xmark & Inappropriate, Discriminating \\
        \citet{ross2017measuring} & Twitter & 469 & \xmark& HS (scale 1-6) \\
        GermEval \citeyear{wiegand2018overview} & Twitter & 8,541 & \xmark & Abuse, Insult, Profanity \\
        GermEval \citeyear{struss2019overview} & Twitter & 7,025 & \xmark & Explicit / Implicit offense\\
        HASOC (\citeauthor{mandl2019overview}) & Twitter, FB & 4,669 & \xmark & HS, Offensive, Profane \\
        \citet{assenmacher2021texttt} & RP & 85,000 & \xmark & Insult, Profane, Threat, Racism \\
        GermEval \citeyear{risch2021overview} & FB & 4,188 & \xmark & Insult, Discrimination, Vulgarity \\
        DeTox \citep{demus-etal-2022-comprehensive} & Twitter & 10,278 & \xmark & 10 target classes for HS\\
        Multilingual HateCheck \citeyear{rottger-etal-2022-multilingual} & Synthetic & 3,645 &  \xmark & Abuse targeted at individuals \\
        GAHD (\citeauthor{goldzycher2024improving}) & Adversarial & 10,996 & \xmark & HS\\
        GERMS-AT (\citeauthor{krenn2024germs}) & DerStandard & 8000 & \xmark & Sexist / Misogynous (scale 0-4)\\
        GerDISDetect (\citeauthor{schutz-etal-2024-gerdisdetect-german})& Media outlets & 1,890 & \xmark & 11 target classes for offense\\
        AustroTox (ours) & DerStandard & 4,562 & \cmark & Spans of Targets and Vulgarities \\
        \bottomrule
    \end{tabularx}
    \caption{Existing German datasets related to offensiveness classification and selected annotations. \textit{HS} stands for hate speech, \textit{FB} for Facebook, and \textit{RP} for Rheinische Post.}
    \label{table:refs}
\end{table*}

For determining the harmfulness of an offensive statement, its target is decisive \citep{bormann2022perceptions, hawkins2023race}.
Perceptions of targets of offensive comments are subject to change based on individual, contextual, cultural and intersectional factors \citep{hawkins2023race, jiang2021understanding, shahid2023decolonizing} making it, therefore, crucial to effectively identify emerging targets of such statements. Figure \ref{fig:examples} depicts examples highlighting the importance of the target of an offensive statement in determining its severity.
In order to study the detection capabilities of language models in an Austrian cultural and linguistic context, we create a corpus of Austrian German comments. 
Our main contributions are: 
\begin{enumerate}
    \item 4,562 user comments from a newspaper discussion forum in Austrian German annotated for offensiveness\footnote{As there are no generally accepted definitions nor distinctions for \textit{abusiveness}, \textit{offensiveness} and \textit{toxicity} \citep{pachinger-etal-2023-toward}, we use these terms interchangeably.} with the article title used as context. The majority of posts is annotated by five annotators. We additionally publish the disaggregated binary offensiveness annotations. 
    \item Annotated spans in comments comprising targeted individuals, groups or other entities by offensive statements, and vulgarities. 
    \item An evaluation of fine-tuned smaller language models and large language models in a zero- and five-shot scenario. 
\end{enumerate}

\section{Related Work}
Research focused on identifying spans within offensive statements is primarily focused on English user comments. Examples of annotated spans in English comments are the targets of offensive statements \citep{zampieri-etal-2023-target}, the spans contributing to the offensiveness label \citep{mathew2021hatexplain, pavlopoulos-etal-2021-semeval}, and the spans comprising a violation of a moderation policy \citep{calabrese2022explainable}. 

We list all public German datasets covering tasks related to offensiveness detection in Table \ref{table:refs}. All German datasets containing labels related to offensiveness except for the One Million Posts and the GerMSDetect dataset focus on different varieties of German from Austrian German. 
AustroTox contains the same definitions for annotating vulgarities as GermEval \cite{risch2021overview}, this dataset contains annotations of vulgar posts. According to their definitions, the classes \textit{Insult} from GermEval \cite{wiegand2018overview}, \textit{Hate Speech} from DeTox and GAHD \citep{demus-etal-2022-comprehensive, goldzycher2024improving}, and \textit{Offense} from HASOC \cite{mandl2019overview} can be merged into the class \textit{Offensive} from AustroTox. This does not imply that the class \textit{Offensive} from AustroTox can be merged into the respective classes as their definition might be more narrow. Additionally, these datasets stem from other sources than AustroTox. 
AustroTox is 
the first German dataset related to offensiveness classification containing annotated spans. 

\section{Dataset Creation}

\paragraph{Data source} We source AustroTox from the Austrian newspaper \textit{DerStandard}\footnote{\url{https://www.derstandard.at/}}, a Viennese daily publication with a left-liberal stance covering domestic and international news and topics such as economy, panorama, web, sport, culture, lifestyle, science, health, education, and family. The DerStandard forum is one of the largest discourse platforms in the German-speaking world. Despite the left-liberal stance of the newspaper, this perspective is not reflective of the forum's community, as DerStandard is actively working on being a low-threshold discussion platform open to everybody. As we focus on the Austrian dialect, this Austrian news media outlet’s comment sections are a suitable sample to draw from. We argue that the forum's expansive community and the diverse range of articles and forums offered on the DerStandard website help towards minimizing bias in AustroTox. Professional moderators ensure the exclusion of hate speech which is illegal
in Austria \citep{austria_hate} in the forum, this results in hardly any hate speech and a focus on offensive speech in the AustroTox dataset. 

\paragraph{Pre-filtering comments} In order to pre-filter potentially toxic comments and comments which are not considered as toxic by existing moderation technologies, we apply stratified sampling  
based on the toxicity score provided by the Perspective API \citep{perspective}. The toxicity score is between $0$ (not toxic) and $1$ (severely toxic). We compute the toxicity score for 123,108 posts. Out of these posts, 873 exhibit a toxicity score between 0.9 and 1. We add these comments to the data to be annotated. 
Furthermore, we create the following strata defined by the toxicity score: 0-0.3, 0.3-0.5, 0.5-0.7, 0.7-0.9. Then, we randomly sample comments from each stratum. We use the following proportions for the counts of comments from the different strata: 9 : 9 : 9 : 11. 

AustroTox encompasses responses to 532 articles or discussion forums on any topics covered by DerStandard. The comments were posted between November 4, 2021, and November 10, 2021. The articles and forums where the comments appear stem from a broader time period. 

\paragraph{Annotation campaign} We conduct the annotation 
with participation from master's students specializing in Data Science and undergraduate students majoring in Linguistics, as an integral component of their academic curriculum. 30\% of the annotators are registered as female through the courses registration platform, which does not necessarily mean that they self-identify as female. The majority of the annotators are Austrian and between 19 and 26 years old, annotators are required to have at least a German level of C1. The vast majority of annotators speak German as a native language. 
Ethical considerations pertaining to the annotation task are expounded upon in the Ethics Statement (Section \ref{sec:Ethics}).

The title of the article under which the comment was posted is taken into account as context when annotating the comment. While our annotation guidelines (Appendix \ref{A:guidelines}) include numerous examples with the intention of being prescriptive \citep{rottger-etal-2022-two}, it is important to note that due to the low number of comments per annotator and the limited time allocated for training the annotators, the procedure unavoidably incorporates a subjective element. 

We classify each comment as offensive or non-offensive. For non-offensive and offensive comments, we annotate spans in the text comprising vulgarities. Both, offensive and non-offensive posts may contain an unspecified number of vulgarities, as vulgar language can exist separate from offensiveness. For offensive posts, we additionally annotate spans comprising the target of the offensive statement and the type of target (Examples in Figure \ref{fig:examples}).  
If the target is only mentioned via a pronoun, we select the pronoun as the span comprising the target. 

Adopting a definition of \textit{vulgarity} similar to that employed by \citet{risch2021overview}, we define classes and spans as follows:
\textbf{\textit{Offensive}}: An offensive comment includes disparaging statements towards persons, groups of persons or other entities or incites to hate or violence against a person or a group of people.
\textbf{\textit{Not Offensive}}: A non-offensive comment does not include disparaging statements or incites to hate or violence.
\textbf{\textit{Vulgarity:}} Obscene, foul or boorish language that is inappropriate  
for civilized discourse. 
\textbf{\textit{Target Group}}: The target of an offensive post is a group of persons or an individual insulted based on shared group characteristics.
\textbf{\textit{Target Individual}}: The target of an offensive post is a single person not insulted based on shared group characteristics.
\textbf{\textit{Target Other}}: The target of an offensive post is not a person or a group of people. 
 
\paragraph{Data aggregation} Each post is annotated by 2 to 5 annotators, the majority of posts is annotated by 5 annotators.  
We choose an aggregation approach that prioritizes sensitivity, where a comment requires fewer votes to be labeled as offensive compared to the number of votes needed to consider it non-offensive. 
A post is solely annotated as non-offensive if $2 \leq v_{n}$ and $v_{o} \leq \frac{v_{n}}{2}$, where $v_{n}$ and $v_{o}$ denote the votes for the class non-offensive and offensive. The post is labelled as offensive if $2 \leq v_{o}$ and $v_{n} \leq \frac{2}{3}\cdot v_{o}$. Posts that do not meet one of these criteria are discarded. This implies that posts which are labelled as offensive by 3 annotators and as non-offensive by 2 annotators are labelled as offensive while posts which are labelled as offensive by 2 annotators and as non-offensive by 3 annotators are discarded. Spans comprising the different target types are annotated by majority voting of those who labelled the post as offensive. 
Vulgarities are annotated if $2\leq v$ and $v_a\leq v + 2$, where $v$ denotes the number of votes for a span being a vulgarity and $v_a$ denotes the sum of all class votes. Table \ref{tab:size} contains the size of AustroTox. 

\begin{table}[ht]
  \centering
  \begin{tabular}{c r r r}
    \toprule    
    & \multicolumn{1}{c}{\textbf{Not Off}} & \multicolumn{1}{c}{\textbf{Off}} 
    \\
    \midrule
    Total & 2,744 & 1,818\\
    \midrule
    Not Vulgar & 2,307 & 712\\
    Vulgar & 437 & 1,106\\
    \midrule  
    No Target & & 34\\
    Target Group & & 869 & \\
    Target Individual & & 572 & \\
    Target Other & & 275 & \\
    More Target Types & & 68\\
    \bottomrule
  \end{tabular}
  \caption{The size of AustroTox. The fine-grained classes are determined by types of spans contained in a comment. Off stands for offensive.}
  \label{tab:size}
\end{table} 

\begin{table*}[ht]
  \centering
  \begin{tabular}{c l c c c c c c c}
  \toprule  
    &
    & 
    & \multicolumn{2}{c}{\textbf{Offensive}} 
    & \multicolumn{2}{c}{\textbf{Vulgar}} 
    & \multicolumn{2}{c}{\textbf{Target}} 
    \\
    &
    & 
    & \multicolumn{2}{c}{Post-level}
    & \multicolumn{2}{c}{Token-level}
    & \multicolumn{2}{c}{Token-level}
    \\
    &
    & \multicolumn{1}{c}{Params} 
    & \multicolumn{1}{c}{Binary}
    & \multicolumn{1}{c}{Macro}
    & \multicolumn{1}{c}{Binary}
    & \multicolumn{1}{c}{Macro}
    & \multicolumn{1}{c}{Micro}
    & \multicolumn{1}{c}{Macro}
    \\
    \midrule \midrule
BERT$_{de}$ & & \multirow{2}{*}{110M} &$.69 \pm .02$&$.74 \pm .02$&$.68 \pm .03$&$.84 \pm .01$&$.22 \pm .04$&$.56 \pm .03 $\\
Bert-dbm & &&$.71 \pm .02$&$.75 \pm .02$&$.69 \pm .02$&$.84 \pm .01$&$.23 \pm .02$&$.57 \pm .03 $\\
GBERT & \multirow{ 2}{*}{Base} &  \multirow{2}{*}{110M}&$.72 \pm .02$&$.76 \pm .01$&$.69 \pm .04$&$.84 \pm .02$&$.23 \pm .02$&$.57 \pm .02 $\\
Gelectra &  &&$.50 \pm .25$&$.57 \pm .12$&$.69 \pm .03$&$\textbf{.85} \pm .01$&$\textbf{.24} \pm .02$&$\textbf{.58} \pm .02 $\\
GBERT & Large & 337M &$.73 \pm .02$&$.75 \pm .03$&$\textbf{.71} \pm .03$&$\textbf{.85} \pm .01$&$.21 \pm .12$&$.53 \pm .16$\\
\midrule
\multirow{2}{*}{LeoLM} & 0-Shot & \multirow{2}{*}{7B} & $.61 \pm 02$ & $.54 \pm 03$ & - & - & - & -  \\
& 5-Shot &  & $.52 \pm 02$ & $.57 \pm 02$ & - & - & - & -  \\
\multirow{2}{*}{Mistral} & 0-Shot & \multirow{2}{*}{7.24B} & $.30 \pm 02$ & $.53 \pm 01$ & - & - & - & -  \\
& 5-Shot &  & $.55 \pm 03$ & $.64 \pm 02$ & - & - & - & -  \\
\multirow{2}{*}{GPT 3.5} & 0-Shot & \multirow{2}{*}{-} & $.68 \pm 01$ & $.64 \pm 02$ & $.40 \pm 02$ & $.69 \pm 01$ & $.17 \pm 01$ & $.52 \pm 01$ \\
& 5-Shot &  & $.72 \pm 01$ & $.72 \pm 02$ & $.43 \pm 02$ & $.70 \pm 01$ & $.20 \pm 01$ & $.55 \pm 01$ \\
\multirow{2}{*}{GPT 4} & 0-Shot & \multirow{2}{*}{-} & $.70 \pm 02$ & $.77 \pm 02$ & $.36 \pm 04$ & $.67 \pm 02$ & $.20 \pm 02$ & $.55 \pm 03$ \\
& 5-Shot &  & $\textbf{.76} \pm 03$ & $\textbf{.81} \pm 02$ & $.41 \pm 02$ & $.70 \pm 01$ & $.22 \pm 03$ & $\textbf{.58} \pm 03$ \\
    \bottomrule
  \end{tabular}
  \caption{Mean $F_1$ scores and standard deviations of ten-fold cross-validation of the classification tasks. We compute the Micro F1 by adding up True and False Positives and False Negatives for the three target classes.}\label{tab:eval_outcomes}
\end{table*}

\paragraph{Inter Annotator Agreement} After curating 390 posts with implausible span annotations (e.g. offensive but no target), we report a Krippendorff's Alpha of $\alpha = 0.49$  on the binary offensiveness classification, which is comparable to related work using crowdsourcing: \citet{sap-etal-2020-social} report $0.51$ and \citet{wulczyn2017ex} report  $0.45$. An $\alpha$ of $0.5$ is between random annotation ($\alpha = 0$) and full agreement ($\alpha = 1$). In a prescriptive annotation paradigm \citep{rottger-etal-2022-two}, tentative conclusions are still acceptable with $\alpha \geq 0.667$ \cite{krippendorff2018content}. While our annotation guidelines include numerous examples, it is important to note that due to the low number of comments per annotator and the limited time allocated for training the annotators, the procedure incorporates a subjective element. Care should be taken when aggregating data in cases of moderate agreement. We argue that our aggregation approach prioritizing sensitivity provides a larger decision boundary. 

\paragraph{Cross-validation splits}
We make AustroTox available with predetermined splits for cross-validation stratified using fine-grained classes determined by the label of the post and the types of spans it contains. The splits consist of a ratio of about 80\% for training, 10\% for development, and 10\% for testing. Appendix \ref{A:dataset_creation} contains more details on the dataset creation. 

\section{Experiments}

\paragraph{Fine-tuned language models}
We fine-tune and evaluate German BERT and Electra models \citep{chan-etal-2020-germans} (Table \ref{tab:eval_outcomes}, Appendix \ref{a:models}).
We define three tasks: Binary offensiveness classification as sequence classification, vulgarity extraction as token classification and target extraction as token classification task.
For offensiveness classification, we concatenate the article title given as context and the comment as input for the models: \verb|article title: <article title>| \verb|\t comment: <comment>|. 

\paragraph{Prompted LLMs}
We additionally evaluate the class and span detection capabilities of not fine-tuned LLMs. We use the following large language models for our experiments: GPT 3.5\footnote{\url{https://platform.openai.com/docs/models/gpt-3-5}} (\textit{gpt-3.5-turbo-1106})
\citep{ouyang2022training},  GPT 4 \footnote{\url{https://platform.openai.com/docs/models/gpt-4-and-gpt-4-turbo}} (\textit{gpt-4-1106-preview}) \citep{achiam2023gpt}, LeoLM 7B Chat \footnote{\url{https://huggingface.co/LeoLM/leo-hessianai-7b-chat}}, and Mistral \footnote{\url{https://huggingface.co/mistralai/Mistral-7B-v0.1}} 
\cite{jiang2023mistral}. We avaluate them in a zero-shot and five-shot scenario (Table \ref{tab:eval_outcomes}, Appendix \ref{a:models}). 

For the LLM evaluation, we distinguish between multitask prediction (predicting offensiveness, vulgarities and targets) and offensiveness-only classification. We create prompts that contain an offensiveness definition, article title, and the post to be classified. In the five-shot scenario, we additionally provide five titles and posts with labels that are randomly sampled from the training set for each prediction. We require the LLM to respond in JSON\footnote{
\url{https://platform.openai.com/docs/guides/text-generation/json-mode}}.  
Preliminary experiments showed that only GPT-3.5 and GPT-4 were able to produce consistently valid JSON responses. We thus only evaluate these two models in the multi-task setup. 
For LeoLM and Mistral, we adjust the prompt, requiring them to respond with only \textit{0} or \textit{1}, 
and define the token with the higher logit as the model's prediction. To ensure comparability for the token-level classification tasks, we tokenize the spans generated by the GPT-models with the GBERT tokenizer.

\paragraph{Evaluation outcomes}
Table \ref{tab:eval_outcomes} contains the evaluation outcomes.
The proprietary LLMs outperform the open-source fine-tuned models in binary offensiveness classification. 
We attribute the superiority of the fine-tuned models in the vulgarity detection task to the lexical nature of the vulgarity detection task. Notably, the dataset features vulgarities in Austrian dialect that are rarely encountered elsewhere. There are 437 non-offensive but vulgar comments in AustroTox. Being able to detect vulgarities can help with debiasing vulgar False Positives and vulgar False Negatives. In especially, the results suggest that marking vulgarities using fine-tuned models and then classifying the comment with marked vulgarities using GPT-4 leads to an improvement of GPT-4’s performance. Even dictionary-based detection of vulgarities might lead to an improvement of GPT-4’s performance and to more explainable results. The span annotations allow for analysis beyond comparing disagreements with binary gold labels. 

The micro $F_1$ on the targets for the four-class target classification is generally low due to a high prevalence of the negative class. 
The fine-tuned models slightly outperform the LLMs at detecting the targets of offensive statements. 

\section{Conclusion}
We presented AustroTox, a dataset comprising user comments in Austrian German, annotated for offensiveness. 
We annotated 
spans within the comments comprising targeted individuals, groups, or other entities through offensive statements and 
spans comprising vulgarities. 
An evaluation on our dataset indicates that the smaller language models we fine-tuned and tested excel in detecting vulgar dialect, whereas the LLMs we tested demonstrate superior performance in identifying offensiveness within AustroTox. 

\section*{Acknowledgements}
Pia and Anna are funded by the Vienna Science and Technology Fund (WWTF) [10.47379/ICT20015]. Janis is funded by the  University of Zurich Research Priority Program project \textit{Digital Religion(s)}. We would like to thank the students participating in the annotation campaign. Their dedication and effort have been invaluable to the success of this project. Furthermore, we would like to thank Rebekah Wegener who helped with the annotation campaign. We would like to extend our gratitude to DerStandard for sharing their data, thereby contributing significantly to the advancement of semi-automated content moderation. Lastly, the financial support by 
the Christian Doppler Research Association is gratefully acknowledged.

\section*{Ethics Statement}\label{sec:Ethics}

\paragraph{Annotators’ Risks}
The repeated exposure of annotators to offensive content carries risks. Therefore, the annotation campaign was reviewed by the ethics committee of our institution. In the course of our work, the annotators engaged in the annotation of comments for a duration of approximately 1.5 to 3 hours. It is noteworthy that the dataset contained a higher proportion of offensive comments than the typical distribution in a user forum. 
The comments were sourced from a publicly accessible, moderated forum by DerStandard, ensuring that none of them could be categorized as illegal under Austrian law \citep{austria_hate}. To mitigate potential distress, the annotators were explicitly informed that they had the option to cease annotation if they felt overwhelmed by the task without facing consequences (Appendix \ref{A:guidelines_generalities}). 

\paragraph{Compensation for Annotators}
Participants in the annotation campaign are predominantly Master students engaged in courses focused on introductory language technology, data annotation, and natural language processing. We consider hands-on experience in annotation tasks to be highly valuable for these students, as it equips them with the necessary skills to potentially design annotation tasks in the future and to be aware of potential pitfalls and difficulties of such tasks. Moreover, we are confident that the expected workload of 1.5 to 3 hours is suitable for the participants. The annotators were informed about the publication of the data and as data annotation is a tedious task, they received a comprehensive compensation through course credits for their efforts.

\paragraph{Risks of Publication of the Data}
There is a potential for exploitation of our results to generate offensive online content that may elude contemporary detection systems. We believe that these risks are manageable when weighed against the improvement of the detection of offensive statements facilitated by AustroTox. 
We urge researchers and practitioners to uphold the privacy of the authors of posts 
when working with the data. And while the data is publicly available on the website of DerStandard, in order to preserve the privacy of users, we replace mentions of users and URLs with special tokens in AustroTox. DerStandard agrees to the publication of the data. Regarding copyright concerns, simple comments in online forums are usually not covered by copyright law §Section 1 UrhG (Austrian Copyright Act).

\section*{Limitations}

\paragraph{Time Span and Range of Topics}
The dataset comprises comments from November 4, 2021, to November 10, 2021 and therefore consists of a higher proportion of COVID-19 related topics. However, we source comments appearing in over 532 varying articles and discussion forums, ensuring diversity in topics in the dataset. The articles and forums where the comments appear stem from a broader time period. Thus, posts in our dataset refer to more events than the ones covered in between November 4, 2021, to November 10, 2021.

\paragraph{Subjectivity}
In the realm of human data annotation for tasks related to sentiment, a degree of subjectivity exists. Due to the small load of comments per annotator, a large pool of annotators, and limited time allocated for training the annotators, this subjective element is reflected by the dataset and is learnt by the models during the training process. We posit that the token-level annotations included in AustroTox elevate the quality of annotation by providing clearer guidance to the annotators. 
Furthermore, we choose an aggregation approach that prioritizes sensitivity, where a comment requires fewer votes to be labeled as offensive compared to the number of votes needed to consider it non-offensive. We posit that this method mitigates lower agreement levels. Additionally, upon publication, we publish the disaggregated annotations for binary offensiveness.

\paragraph{Limitations of Experiments}
Our computational experiments on the dataset are not yet exhaustive, as certain models, notably larger encoder-only Transformer models trained on German data (like GELECTRA-large), may outperform the encoder-only Transformer models examined in our study. We did not conduct hyperparameter-search which would further improve the outcomes of the evaluation. Moreover, it is essential to include more thorough testing of open-source LLMs alongside the GPT models. Our future aim is to deliver a more comprehensive evaluation on the dataset, enabling a nuanced consideration of factors influencing the performance difference of fine-tuning and zero-shot classification on larger models.

\fancyhf{} 
\bibliography{anthology,custom}

\appendix

\section{Annotation Guidelines
}\label{A:guidelines}
We depict the annotation guidelines used for the dataset creation. 
For the paper, we translate the example comments (except for vulgarities in dialect) with DeepL and correct the translations manually where the overall meaning of the comment is not preserved, note that we do not try to make the comment sound as if it was written by a native English speaker. We add the original comments as footnotes. Vulgar passages might not be vulgar in the English version, nevertheless, we indicate the annotation of the original German comment. Following \citet{nozza-hovy-2023-state}, we obfuscate vulgarities. Originally, we annotated insults and incites to hate or violence. Nevertheless, by majority voting only 60 comments were labelled as \textit{Incite to hate or violence}, therefore, we merge the two classes into an \textit{Offensiveness} class. 

\subsection{Generalities}\label{A:guidelines_generalities}
\textbf{Your mental health}\hspace{6pt} If the comments are disturbing for you, please stop annotating and contact us. This would of course not affect your grade, we would find a solution together on how to grade you.\\
\textbf{Classes}\hspace{6pt} There are three exclusive classes. You select a class for each comment. See Section \textit{Classes} for details.\\
\textbf{Spans}\hspace{6pt} There are four spans. The spans are used for tagging passages. In order to do so, mark the text you would like to tag and select one of the four tags for the spans. See Section \textit{Spans} for details.\\
\textbf{Title of article}\hspace{8pt} You can see the title of the article which was commented at the right side of the page. Please take it into account when classifying and tagging the comment.\\
\textbf{Subjectivity}\hspace{6pt} It is often hard to classify a comment with one of these classes as there are many nuances of \textit{insults} and \textit{incites to hate and violence} (such as for example irony). If you are not sure about how to label a comment, choose the most reasonable option to you, sometimes there is no right or wrong.\\
\textbf{German level}\hspace{6pt} If you realize that the level of German in these comments is too hard for you or there is too much dialect in them, don't hesitate to contact us, we can easily assign you the English task and you annotate the remaining comments in English.

\subsection{Class Insult}
An insult includes disparaging statements towards persons or groups of persons as well as towards other entities. Insults pursue the recognisable goal of disparaging the addressee or the object of reference. \textbf{Examples:}
\begin{itemize}
    \item \textit{Arguments? If this \underline{group} 
    of \underline{vaccination refusers} could be reached with reason and arguments, our situation would be different.} \footnote{\textit{Argumente?. Wenn diese \underline{Bagg*ge} von \underline{Impfverweigerern} mit Vernunft und Argumenten erreichbar wäre, sähe unsere Lage anders aus.}} 
    \begin{itemize}
        \item \tag{Class} Insult
        \item \tag{Span} Target Group: \textit{vaccination refusers}
        \item \tag{Span} Vulgarity: \textit{group}
    \end{itemize}
    \item \textit{\underline{They}'re definitely \underline{crazy}... I'm not allowed to go to the Inn, I'm not allowed to ski, I'm supposed to make up for their losses with my taxes.   You're \underline{out of your mind}!  And no thanks, you can save yourselves the \underline{st*pid} slogans, well then just get vaccinated.} \footnote{\textit{\underline{Die} \underline{sp*nnen} doch endgültig.. Ich darf nicht ins Gasthaus, ich darf nicht Skifahren, soll deren Ausfälle mit meinen Steuern ausgleichen.   Ihr \underline{h*bt sie doch nicht alle}!  Und nein danke, die \underline{besch**erten} Sprüche na dann geh halt impfen, könnt ihr euch gleich sparen.}} 
    \begin{itemize}
        \item \tag{Class} Insult
        \item \tag{Span} Target Group: \textit{They}
        \item \tag{Span} Vulgarity: \textit{crazy}
        \item \tag{Span} Vulgarity: \textit{out of your mind}
        \item \tag{Span} Vulgarity: \textit{st*pid}
    \end{itemize}
    \item \textit{LOL Our \underline{vaccine refusers in the football club} are all pissed off because they can't go to the Christmas party now XD \underline{g**ns.}}\footnote{\textit{Lol Unsere \underline{impfverweigerer im Fußball Verein} sein voll angfressn weil's jetzt ned auf die Weihnachtsfeier gehen können XD \underline{D*llos}}} 
    \begin{itemize}
        \item \tag{Class} Insult
        \item \tag{Span} Target Group: \textit{vaccination refusers in the football club}
        \item \tag{Span} Vulgarity: \textit{g**ns.}
    \end{itemize}
    \item \textit{Should i applaud to that \underline{sh*t?}} \footnote{\textit{Soll i zu dem \underline{Sch**ß} auch noch applaudieren?}} 
    \begin{itemize}
        \item \tag{Class} Insult
        \item \tag{Span} Target Other: \textit{sh*t}
        \item \tag{Span} Vulgarity: \textit{sh*t}
    \end{itemize}
\end{itemize}

\subsection{Class Incite to Hate or Violence}
An incite to hate or violence against a person or a group of people. It is hard to draw the line between \textit{insults} and \textit{incites to hate}, as insults always somewhat incite hate. Try to decide for yourself what you actually consider hate and what is more of an insult. Insults are usually less severe. If a comment includes an insult and an incite to hate or violence, please choose the class \textit{Incite to hate or violence}. \textbf{Examples:}
\begin{itemize}
   \item \textit{You just have to stand up to them. Yesterday you could see \underline{migrants} with tools trying to cut the fence. You simply have to drive over it.} \footnote{\textit{Da muss man einfach hart dagegen halten. Gestern konnte man ja \underline{Migranten} mit Werkzeug sehen, die den Zaun zerschneiden wollten. Da gehört einfach drübergefahren.}}
    \begin{itemize}
        \item \tag{Class} Incite to hate or violence
        \item \tag{Span} Target Group: \textit{migrants}
    \end{itemize}
   \item \textit{\underline{Refugees} should face the squad!}
    \begin{itemize}
        \item \tag{Class} Incite to hate or violence
        \item \tag{Span} Target Group: \textit{Refugees}
    \end{itemize}
   \item \textit{All \underline{Austrian people} are dirty!} 
    \begin{itemize}
        \item \tag{Class} Incite to hate or violence
        \item \tag{Span} Target Group: \textit{Austrian people}
    \end{itemize}
\end{itemize}

\subsection{Class None}
None of the above classes. \textbf{Examples:}
\begin{itemize}
    \item \textit{KC-GB is already a pretty weak game... Will Mahomes find his old strength at some point this season? I don't think he will this year...}\footnote{\textit{KC-GB ist schon ein ziemlich schwaches Spiel...Ob Mahomes irgendwann in dieser Saison zu alter Stärke findet? Glaub heuer wird des nix mehr...}}
    \begin{itemize}
        \item \tag{Class} None
    \end{itemize}
    \item \textit{they think something will happen with these rules. And the Burgenlanders look \underline{st*pid} again. That's how "motivation" works.} \footnote{\textit{die glauben mit diesen Regeln passiert irgendwas. Und die Burgenländer schauen wieder \underline{bl*d} aus. So funktioniert \"Motivation\"}}
    \begin{itemize}
        \item \tag{Class} None
        \item \tag{Span} Vulgarity: \textit{st*pid}
    \end{itemize}
    \item \textit{yes anyway, the Ministry of Finance obviously paid for it ... \underline{wtf}} 
    \footnote{\textit{ja eh hat ja offensichtlich das finanzministerium bezahlt ... \underline{wtf}}}
    \begin{itemize}
        \item \tag{Class} None
        \item \tag{Span} Vulgarity: \textit{wtf}
    \end{itemize}
    \item \textit{Oh, oh, why are we \underline{sh*t?}} \footnote{\textit{Oh, oh, warum samma \underline{sch**ße}?}}
    \begin{itemize}
        \item \tag{Class} None
        \item \tag{Span} Vulgarity: \textit{sh*t}
    \end{itemize}
    \item \textit{She's upset about 2G but what does she suggest? Should we just let people die?} \footnote{\textit{Sie regt sich über 2G auf aber was schlägt sie denn vor? Sollen wir die Leute einfach verrecken lassen? }}
\end{itemize}

\subsection{Spans}
In contrast to the classes, comments can have zero, one, or multiple spans. Insults and incitements to hate or violence are targeted at an individual person, a group of persons, or something else, such as, for example, democracy (\textit{Target Other}). Therefore, for comments classified as \textit{Insult} or \textit{Incite to hate or violence}, you tag at least one person or thing as \textit{Target Individual}, \textit{Target Group}, or \textit{Target Other}. For the class \textit{None}, you \underline{don't} tag \textit{Target Individual}, \textit{Target Group}, or \textit{Target Other}. Vulgar passages may be found in comments of all three classes.

\subsection{Span Vulgarity}
Obscene, foul or boorish language that is inappropriate or improper for civilised discourse. 
\begin{itemize}
    \item \tag{Example} \textit{\textit{What the F*****ck}... Magnificent Interception.} \footnote{\textit{\underline{What the F*****ck}... Grandiose Interception.}}
    \begin{itemize}
        \item \tag{Class} None
        \item \tag{Span} Vulgarity: \textit{What the F*****ck}
    \end{itemize}       
\end{itemize} 
Examples of vulgar expressions:  \textit{P*mmel, M*st, schw*chsinnig, Sh*tty, Gsch*ssene, v*rtrottelten, D*mmling-Däumlinge, WTF, Schn*sel, v*rsaut, W*ppla, K*ffer, zum T**fel, Tr*ttel, D*pp, D*dl, Sch**xx, d*mn, Hosensch**sser-Nerds, N*zipack, Gsch*ssene, Ges*ndel, verbl*dete, Schw*chköpfe, sch**ßen gehen, Schattenschw*nzler, p*ppn, P**fkinesen}

\subsection{Span Target Individual}
The target of an insult or an incite to hate or violence is a single person not insulted based on shared group characteristics.
\begin{itemize}
    \item \tag{Example} \textit{f*ck you, Max Mustermann}
    \begin{itemize}
        \item \tag{Class} Insult
        \item \tag{Span} Target Individual: \textit{Max Mustermann}
        \item \tag{Span} Vulgarity: \textit{f*ck}
    \end{itemize}
\end{itemize}

\subsection{Span Target Group} 
The target of an insult or an incite to hate or violence is a group of persons or an or an individual insulted based on shared group characteristics.
    \begin{itemize}
\item \tag{Example} \textit{You have to treat \underline{id**ts} like \underline{id**ts}! Therefore, of course, lockdown to save the economy! } \footnote{\textit{Mit \underline{Id**ten} muss man wie \underline{Id**ten} umgehen! Daher selbstverständlich Lockdown, um die Wirtschaft zu retten!}}
        \begin{itemize}
            \item \tag{Class} Insult
            \item \tag{Span} Target Group: \textit{id**ts}
            \item \tag{Span} Vulgarity: \textit{id**ts}

        \end{itemize}
    \end{itemize}

\subsection{Span Target Other} The target of an insult or an incite is not a person or a group of people. \textbf{Examples:}
    \begin{itemize}
        \item \textit{Please what kind of \underline{st*pid} \underline{regulation} is this USRTOK} \footnote{\textit{Bitte was ist denn das für eine \underline{d*mme} \underline{Regelung} USRTOK}}
        \begin{itemize}
            \item \tag{Class} Insult
            \item \tag{Span} Target Other: \textit{regulation}
            \item \tag{Span} Vulgarity: \textit{st*pid}     
        \end{itemize}
        \item \textit{yes please \underline{f*ck around a little longer}, finally make some decisions, the actions of our \underline{government} are a disaster} \footnote{\textit{ja bitte \underline{sch**ßts noch a bisserl länger um}, endlich mal Entscheidungen treffen, das Vorgehen unserer \underline{Regierung} ist ein Desaster}}
        \begin{itemize}
            \item \tag{Class} Insult
            \item \tag{Span} Target Other: \textit{government}
            \item \tag{Span} Vulgarity: \textit{f*ck around a little longer}
        \end{itemize}
    \end{itemize}

\section{Details on Dataset Creation}\label{A:dataset_creation}
\begin{figure}[]
\includegraphics[width=\linewidth, height=3cm]{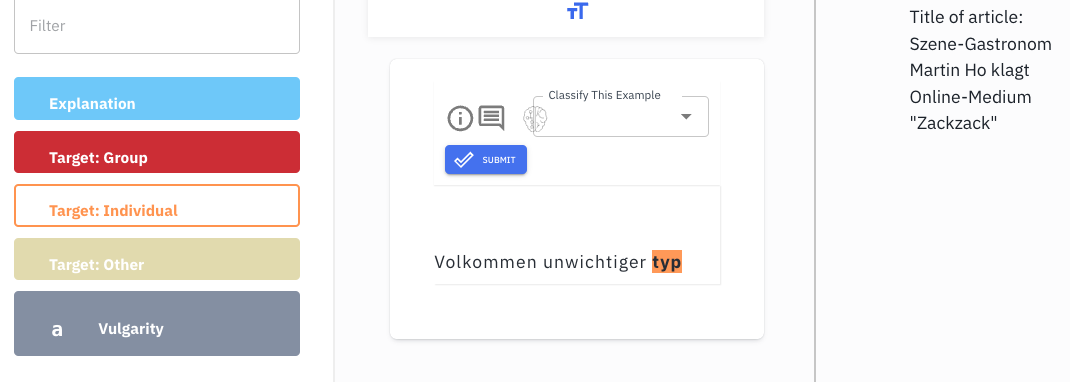}
\caption{The LightTag interface}
\label{fig:lighttag}
\end{figure}

We exclusively select original comments for the dataset while excluding responses to other comments. 
We use the annotation tool LightTag \footnote{\url{https://www.lighttag.io/}} (Figure \ref{fig:lighttag}). Each annotator undertakes the task of annotating a volume ranging from 200 to 300 comments. 
We use \citeposs{castro-2017-fast-krippendorff} implementation for the Krippendorff's Alpha.
Mentions of users and URLs are replaced with \textit{USRTOK} and \textit{URLTOK}.

\section{Details on Experiments}
We utilize OpenAI Copilot for code implementation.

\paragraph{Models}\label{a:models}
We use the following models (with licenses in parenthesis) suitable for fine-tuning for downstream tasks or for few-shot classification for our experiments:
\begin{enumerate}
    \item BERT$_{de}$ cased 
    \footnote{\url{https://huggingface.co/bert-base-german-cased}} (MIT)
    \item BERT-dbmdz\footnote{\url{https://huggingface.co/dbmdz/bert-base-german-cased}} 
    (MIT)
    \item GELECTRA 
    \footnote{\url{https://huggingface.co/deepset/gelectra-base}} (MIT)
    \item GBERT-base  
    \footnote{\url{https://huggingface.co/deepset/gbert-base}} (MIT)
    \item GBERT-large 
    \footnote{\url{https://huggingface.co/deepset/gbert-large}} (MIT)
    \item LeoLM 7B Chat \footnote{\url{https://huggingface.co/LeoLM/leo-hessianai-7b-chat}} (LLAMA 2 COMMUNITY LICENSE AGREEMENT)
    \item Mistral \footnote{\url{https://huggingface.co/mistralai/Mistral-7B-v0.1}} (Apache 2.0)
    \item ChatGPT -- \textit{gpt-3.5-turbo-1106} \footnote{\url{https://platform.openai.com/docs/models/gpt-3-5}} 
    \item GPT4 -- \textit{gpt-4-1106-preview} \footnote{\url{https://platform.openai.com/docs/models/gpt-4-and-gpt-4-turbo}}
\end{enumerate}

\paragraph{Fine-Tuning}
We use the Transformer-models' implementation from the huggingface \texttt{Transformers} library with the default values (huggingface-hub version 0.17.3, Transformers version 4.34.0). 
For all models, we use a learning rate of $5e^{-5}$, weight decay $0.01$ with 200 warm-up steps.
We use a per-device train batch size of 8 examples. 
We train the models for a maximum of 10 epochs, with early early stopping at a patience of 3 epochs. We keep the model with the best Binary or Micro F1 score. 
We use four Nvidia GTX 1080 TI GPUs with 11GB RAM each to train each model. 
Training all offensiveness classification models took about 20 GPU hours in total, whereas vulgarity and target extraction models 16 GPU hours each. 

\begin{figure*}[t]
\includegraphics[width=12cm]{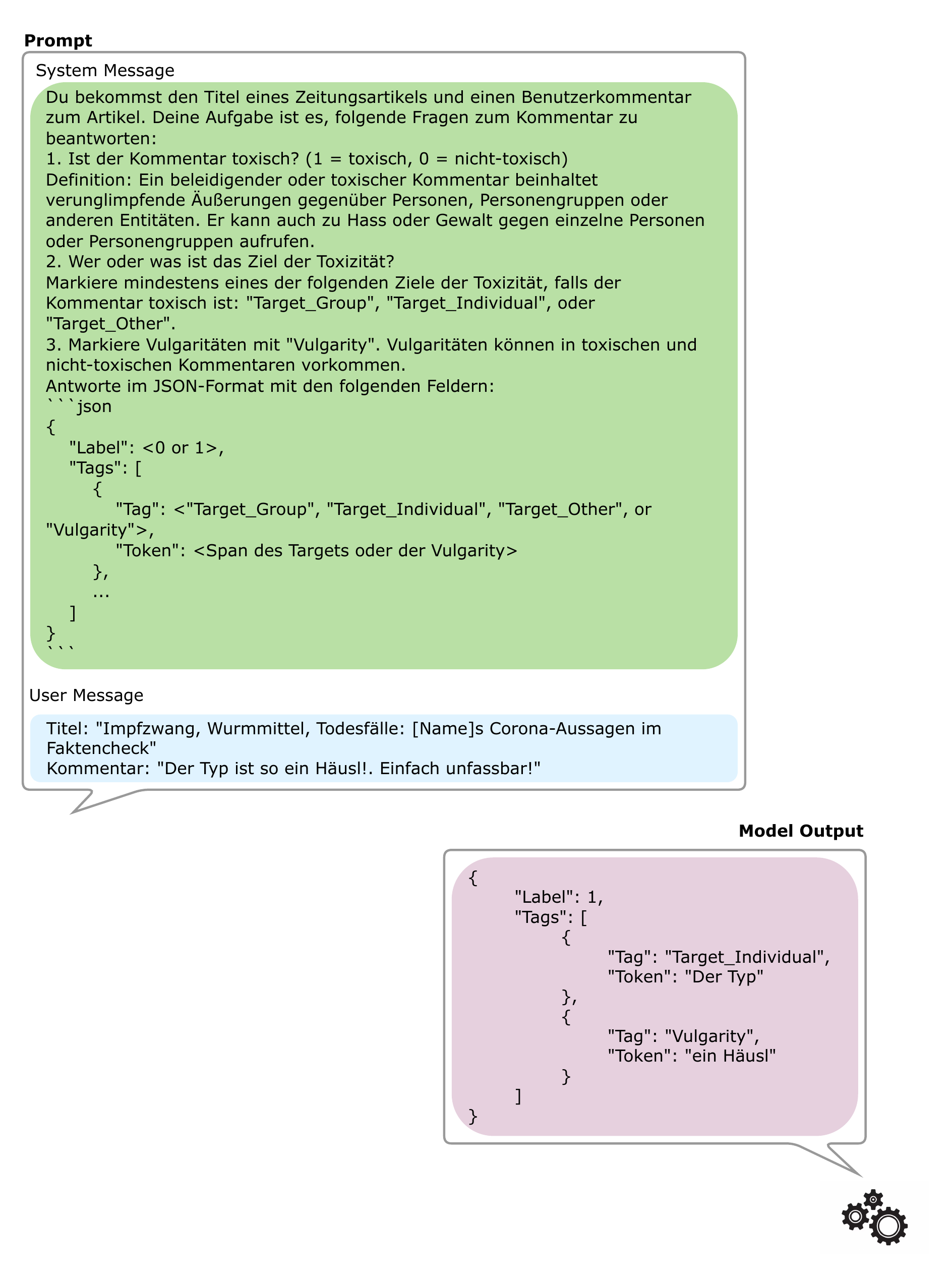}
\caption{The prompt we provide and an example response.}
\label{fig:gpt-prompt}
\end{figure*}

\paragraph{Prompting} \label{appsec:prompting}
Figure \ref{fig:gpt-prompt} contains the zero-shot prompt for the multitask-setup. 
For the Llama 2 based models, we add the Llama-style start and end spans to the prompts.
To evaluate LeoLM and Mistral AI we use a cluster with eight NVIDIA GeForce RTX 3090 GPUs (24 GB RAM per GPU). We estimate two GPU hours per evaluated model.

\paragraph{Evaluation}
We compute the Micro F1 for the target classification using the framework of \citealt{seqeval}.

\end{document}